\documentclass[10pt,twocolumn,letterpaper]{article}

\usepackage{cvpr}
\usepackage{times}
\usepackage{epsfig}
\usepackage{graphicx}
\usepackage{amsmath}
\usepackage{amssymb}
\usepackage{booktabs}
\usepackage{multirow}
\usepackage{array}
\usepackage{color}
\usepackage{xcolor}
\usepackage{bm}
\usepackage{algorithm}
\usepackage{algorithmic}
\usepackage{subfigure}
\usepackage{hyperref}

\begin{document}

\title{Selective Attention-Based Network for Robust Infrared Small Target Detection}

\author{
Yingming Zhang$^{1}$ \quad Wuqi Su$^{1}$ \quad Qing Xiao$^{2*}$ \quad Yonggang Yang$^{2}$ 
\\
$^{1}$School of Software and Internet of Things Engineering, Zhejiang Gongshang University \\
$^{3}$School of Computer Science and Technology, Tiangong University \\
{\tt\small \{zhangyingming, suwuqi\}@zjgsu.edu.cn}
}

\maketitle

\begin{abstract}
Infrared small target detection (IRSTD) plays a pivotal role in a broad spectrum of mission-critical applications, including maritime surveillance, military search and rescue, early warning systems, and precision-guided strikes, all of which demand the precise identification of dim, sub-pixel targets amid highly cluttered infrared backgrounds. Despite significant progress driven by deep learning methods, fundamental challenges persist: infrared small targets occupy extremely limited spatial extents (often only a few pixels), exhibit low signal-to-clutter ratios, and are easily confused with structurally complex backgrounds that frequently induce false alarms. Existing encoder--decoder architectures suffer from two key limitations---an information bottleneck in early convolutional stages that undermines fine-grained target perception, and static skip connections that lack the dynamic adaptability required to discriminate between genuine targets and pseudo-target regions. To address these challenges, we propose SANet, a Selective Attention-based Network built upon the classical U-Net framework and augmented with two novel components: (1)~a \emph{Dual-path Semantic-aware Module} (DSM) that integrates standard convolutions for local spatial detail preservation with pinwheel-shaped convolutions for expanded, direction-sensitive receptive fields, followed by a Convolutional Block Attention Module (CBAM) for fine-grained spatial--channel feature recalibration; and (2)~a \emph{Selective Attention Fusion Module} (SAFM) that replaces conventional static skip connections with a spatially adaptive, learnable weighting mechanism to perform context-aware, cross-scale feature fusion. Extensive experiments on three widely used benchmarks---NUAA-SIRST, IRSTD-1K, and NUDT-SIRST---demonstrate that SANet consistently outperforms fourteen state-of-the-art methods across all evaluation metrics, achieving IoU improvements of 1.93\%, 4.32\%, and 2.21\% over the respective second-best approaches, thereby confirming its strong generalization and practical applicability. 

\end{abstract}
\section{Introduction}
\label{sec:intro}

Infrared small target detection (IRSTD) constitutes one of the most challenging and practically relevant tasks in computer vision and remote sensing. The fundamental goal of IRSTD is to accurately identify and localize extremely dim, spatially compact targets embedded within complex infrared backgrounds~\cite{rawat2024irstd_review, li2024infrared_survey}. The importance of this task cannot be overstated: IRSTD forms the technological backbone of maritime surveillance systems, military search-and-rescue operations, airborne early warning platforms, and precision-guided munitions~\cite{yang2024pbt}. In all of these application domains, the accuracy and reliability of detection results directly impact mission success and personnel safety, making the pursuit of more effective IRSTD algorithms a research priority of enduring significance.

However, IRSTD presents a constellation of unique and formidable challenges that distinguish it from conventional object detection tasks in natural images. First, infrared small targets are characterized by their extremely limited spatial resolution, frequently spanning only a handful of pixels in the image plane, which renders them nearly indistinguishable from noise artifacts and background clutter~\cite{dai2021acm, zhang2022isnet}. Second, these targets typically exhibit a low signal-to-clutter ratio (SCR) and weak thermal contrast against the background, particularly in dynamic environments such as cloud-covered skies, turbulent sea surfaces, and urban heat islands~\cite{sun2023rdian}. Third, infrared backgrounds are structurally complex and highly diverse, containing edges, textures, and bright spots that closely mimic the appearance of genuine targets and frequently induce false alarms~\cite{han2020local_contrast, han2021weighted}. Taken together, these factors conspire to make the target easily obscured by or confused with the surrounding background, rendering IRSTD a persistently challenging problem.

\textbf{Traditional methods.} Historically, IRSTD methods have been categorized into three principal families: filtering-based approaches, low-rank representation methods, and local contrast measures. Filtering-based methods~\cite{deng2021adaptive, bai2010top_hat} operate in the spatial or frequency domain to suppress background interference and have demonstrated satisfactory performance in relatively simple, homogeneous backgrounds. However, their effectiveness degrades markedly in scenarios where the spectral characteristics of targets and backgrounds overlap significantly. Low-rank representation methods~\cite{dai2017reweighted, sun2021subspace, zhang2019pstnn} decompose the infrared image into a low-rank background matrix and a sparse target matrix, achieving effective target enhancement under low-SCR conditions. Nevertheless, their fundamental inability to reliably distinguish sparse noise from genuine targets limits their practical utility, often resulting in elevated false alarm rates. Local contrast methods~\cite{han2020local_contrast, han2021weighted, qin2019fkrw} enhance target saliency by computing intensity differences within local neighborhoods, offering a degree of detection capability for small targets. However, their rigid dependence on fixed window sizes constrains their adaptability to multi-scale targets and varying background conditions.

\textbf{Deep learning methods.} In recent years, deep learning methods have revolutionized the IRSTD landscape, leveraging their powerful feature learning capabilities to achieve substantial improvements in detection accuracy and robustness. The seminal work of Dai~\emph{et al.}~\cite{dai2021acm} introduced the first segmentation-based detection paradigm and proposed an asymmetric contextual modulation (ACM) structure to enhance multi-scale feature fusion within the U-Net framework~\cite{ronneberger2015unet}. Building upon this foundation, Li~\emph{et al.}~\cite{li2023dnanet} developed the dense nested attention network (DNANet) to achieve fine-grained multi-scale feature fusion for infrared small targets. Zhang~\emph{et al.}~\cite{zhang2022isnet} proposed ISNet, which explicitly models the shape characteristics of infrared targets, demonstrating superior performance across multiple benchmarks. More recently, Yang~\emph{et al.}~\cite{yang2024pbt} introduced the progressive background-aware Transformer (PBT), which employs a multi-stage background context refinement mechanism to mitigate feature degradation and error accumulation. Wu~\emph{et al.}~\cite{wu2025l2sknet} proposed learnable local saliency kernels to replace conventional cross-layer feature fusion, guiding the network to capture target-specific salient features more effectively. The application of state space models~\cite{gu2021efficiently, gu2023mamba} to IRSTD has also gained momentum, with MiM-ISTD~\cite{chen2024mim_istd} pioneering the use of the Mamba architecture for this task, and IRMamba~\cite{zhang2025irmamba} further enhancing target detail representation through pixel-difference-aware state updates.

Despite these significant advances, two fundamental limitations persist in current deep learning-based IRSTD methods, motivating the present work:

\begin{enumerate}
    \item \textbf{Information bottleneck in early feature extraction.} Conventional convolutional layers employed in the initial stages of existing networks exhibit insufficient sensitivity to fine-grained spatial information, which is critical for perceiving dim, sub-pixel targets against complex backgrounds~\cite{yu2021nasface}. This perceptual deficiency results in inadequate feature contrast between targets and their surrounding background, directly limiting the network's discriminative capability and, consequently, detection precision and reliability.
    
    \item \textbf{Static and inflexible skip connections.} The static feature fusion mechanisms prevalent in mainstream encoder-decoder architectures, particularly the standard skip connections in U-Net and its variants~\cite{ronneberger2015unet, zhou2018unetpp}, transfer encoder features to the decoder in a spatially uniform manner without considering local saliency or semantic relevance. This fixed-weight fusion strategy fails to dynamically distinguish target regions from background clutter, limiting the model's discriminative capacity under complex background conditions and introducing unnecessary background noise into the decoding pathway.
\end{enumerate}

To overcome these fundamental limitations, we propose \textbf{SANet} (Selective Attention-based Network for Infrared Small Target Detection), a novel architecture that addresses both the feature extraction bottleneck and the skip connection rigidity through two carefully designed modules. First, we introduce the \textbf{Dual-path Semantic-aware Module (DSM)}, which serves as the foundational feature extraction block within the encoder. The DSM integrates a standard convolution branch for preserving fine-grained local spatial details with a pinwheel-shaped convolution branch~\cite{yang2025pinwheel} for expanding the receptive field in a direction-sensitive manner with minimal computational overhead. A Convolutional Block Attention Module (CBAM)~\cite{woo2018cbam} is subsequently applied to perform spatial and channel recalibration, further refining the feature representation. Second, we propose the \textbf{Selective Attention Fusion Module (SAFM)} to replace the static skip connections in the U-Net decoder. The SAFM introduces a spatially adaptive weighting mechanism that dynamically fuses encoder and decoder features based on their spatial saliency, enabling the network to selectively enhance regions of high target probability while suppressing irrelevant background signals.

In summary, the main contributions of this paper are:
\begin{itemize}
    \item We propose SANet, a selective attention-based network for infrared small target detection that substantially improves the model's ability to discriminate between genuine targets and false alarms in complex infrared backgrounds.
    \item We design the Dual-path Semantic-aware Module (DSM), which combines standard and pinwheel-shaped convolutions to achieve expanded receptive fields with effective background noise suppression, augmented by spatial--channel attention for enhanced detail perception.
    \item We introduce the Selective Attention Fusion Module (SAFM), a dynamic alternative to static skip connections that guides selective information transfer through spatially adaptive feature weighting, significantly improving the network's false alarm suppression capability.
    \item Extensive experiments on three public benchmarks demonstrate that SANet outperforms existing state-of-the-art methods across multiple metrics, exhibiting strong detection performance, robustness, and generalization.
\end{itemize}

\section{Related Work}
\label{sec:related}

\subsection{Traditional Infrared Small Target Detection}

Traditional IRSTD methods have historically constituted the foundational approaches in this research domain and continue to serve as important baselines for evaluating modern deep learning alternatives. These classical techniques can be broadly categorized into three principal families, each exploiting different prior knowledge about the characteristics of infrared images.

\textbf{Filtering-based methods} operate by designing spatial-domain or frequency-domain filters to selectively suppress background interference while preserving target signals. Representative approaches include the top-hat transformation~\cite{bai2010top_hat} and its adaptive variants~\cite{deng2021adaptive}, which have demonstrated reliable performance in relatively uniform backgrounds. However, their efficacy deteriorates substantially when the spectral energy distributions of targets and backgrounds exhibit significant overlap, a common occurrence in real-world cluttered infrared scenes.

\textbf{Low-rank representation methods}~\cite{dai2017reweighted, sun2021subspace, zhang2019pstnn, zhang2019nonconvex} formulate the detection task as a matrix decomposition problem, separating the infrared image into a low-rank component (representing the slowly varying background) and a sparse component (representing potential targets). While these methods have achieved notable success in enhancing targets under low-SCR conditions, they suffer from a fundamental ambiguity in distinguishing between sparse noise artifacts and genuine sparse targets, which often leads to unacceptably high false alarm rates in operational scenarios.

\textbf{Local contrast methods}~\cite{han2020local_contrast, han2021weighted, qin2019fkrw} compute intensity differences within local neighborhoods to enhance the saliency of small targets relative to their immediate background. Although conceptually simple and computationally efficient, their rigid dependence on predefined window sizes limits their adaptability to targets of varying scales and to backgrounds with heterogeneous structural complexity.

Despite their historical importance, traditional methods share a common fundamental limitation: their heavy reliance on hand-crafted features and manually tuned hyperparameters constrains their capacity to model the inherent diversity and variability of targets in complex real-world scenes, resulting in detection performance and robustness that are often inadequate for practical deployment.

\subsection{Deep Learning-Based Infrared Small Target Detection}

The advent of deep learning has catalyzed a paradigm shift in IRSTD, enabling data-driven approaches that significantly surpass traditional methods in both accuracy and robustness. The foundational architecture underlying most modern IRSTD methods is the U-Net~\cite{ronneberger2015unet} encoder-decoder framework, which has been extended and enhanced in numerous ways to address the unique challenges of infrared small target detection.

MDvsFA~\cite{wang2019mdvsfa} pioneered the introduction of adversarial learning to IRSTD, explicitly formulating the detection task as a minimax optimization between miss detection and false alarm rates. ACM~\cite{dai2021acm} proposed an asymmetric contextual modulation strategy within the U-Net architecture, fusing high-level semantic information with low-level fine-grained details to improve detection precision. ALCNet~\cite{dai2021alcnet} embedded a parameter-free local contrast measurement module into the network architecture, overcoming the limited receptive field of small convolutional kernels. UIU-Net~\cite{wu2023uiunet} achieved multi-level, multi-scale target representation by nesting a lightweight U-Net within the main U-Net backbone. RPCANet~\cite{wu2024rpcanet} applied deep unfolding of robust principal component analysis to the IRSTD task, combining model-based and data-driven paradigms.

Several recent works have explored attention mechanisms and Transformer architectures to further enhance IRSTD performance. ABC~\cite{pan2023abc} integrated convolutional and linear structures with Transformer modules to enhance both feature extraction and fusion capabilities, effectively suppressing background noise interference. SCTransNet~\cite{yuan2024sctransnet} constructed spatial--channel cross Transformer modules to enhance the separability of targets and background clutter across different semantic levels. AGPCNet~\cite{zhang2023agpcnet} designed attention-guided pyramid context networks for detecting infrared small targets under complex backgrounds.

More recently, the boundaries of IRSTD research have expanded to encompass novel architectural paradigms. GCI-Net~\cite{zhang2024gcinet} leveraged Gaussian curvature information to enhance the geometric features of target regions. CFD-Net~\cite{zhang2025cfd} drew inspiration from computational fluid dynamics, modeling the evolution of infrared features as analogous to particle flow. HDNet~\cite{xu2025hdnet} jointly exploited spatial-domain multi-scale perception and frequency-domain low-frequency suppression. The scale and location sensitivity (SLS) loss proposed by Liu~\emph{et al.}~\cite{liu2024sls} addressed fundamental limitations in existing IoU-based loss functions. State space models have also made significant inroads, with MiM-ISTD~\cite{chen2024mim_istd} and IRMamba~\cite{zhang2025irmamba} demonstrating the potential of the Mamba architecture for modeling long-range dependencies in infrared images. Furthermore, foundation models such as Segment Anything~\cite{kirillov2023segment_anything} have been adapted for IRSTD through specialized frameworks like IRSAM~\cite{zhang2024irsam}.

While these advances have progressively pushed the state of the art, the dual challenge of insufficient fine-grained feature extraction and inflexible multi-scale feature fusion remains insufficiently addressed, motivating the design of our proposed SANet.

\subsection{Attention Mechanisms in Computer Vision}

Attention mechanisms have become indispensable tools in modern computer vision, enabling neural networks to selectively focus on the most informative features while suppressing irrelevant distractions~\cite{vaswani2017attention, hu2018squeeze, woo2018cbam}. The Squeeze-and-Excitation (SE) module~\cite{hu2018squeeze} introduced channel-wise recalibration by modeling inter-channel dependencies through global average pooling and fully connected layers. CBAM~\cite{woo2018cbam} extended this concept by incorporating both spatial and channel attention in a sequential manner, achieving more comprehensive feature refinement. The Transformer architecture~\cite{vaswani2017attention}, originally developed for natural language processing, has been successfully adapted for vision tasks~\cite{dosovitskiy2021vit, liu2021swin}, demonstrating the power of self-attention for capturing long-range dependencies.

In the IRSTD context, attention mechanisms play a particularly critical role because the extreme size disparity between targets and backgrounds demands highly selective feature processing. Our proposed SANet leverages attention mechanisms at two distinct levels: within the DSM for fine-grained feature recalibration, and within the SAFM for spatially adaptive cross-scale feature fusion, thereby addressing the unique challenges of infrared small target detection in a principled and comprehensive manner.

\section{Method}
\label{sec:method}

In this section, we present the detailed architecture of the proposed SANet. We first describe the overall network structure, and then elaborate on the two core modules: the Dual-path Semantic-aware Module (DSM) and the Selective Attention Fusion Module (SAFM).

\subsection{Overall Architecture}

The overall architecture of SANet is illustrated in Figure~\ref{fig:architecture}. The network is built upon the classical U-Net encoder-decoder framework~\cite{ronneberger2015unet} and is augmented with two carefully designed modules to enhance its infrared small target detection capability.

The encoder pathway consists of five encoding stages. Each encoding stage comprises a Dual-path Semantic-aware Module followed by a max-pooling operation that progressively reduces the spatial resolution of the feature maps. The DSM is specifically designed to fuse contextual information from an expanded receptive field while employing CBAM~\cite{woo2018cbam} to selectively enhance target-relevant features and suppress background interference, thereby improving the network's sensitivity to dim, sub-pixel targets.

The decoder pathway comprises four Selective Attention Fusion Modules paired with corresponding upsampling convolutional layers. The SAFM dynamically fuses multi-scale features from the encoder and decoder pathways, adaptively enhancing the response in spatially salient regions to improve the discrimination between genuine small targets and false alarm candidates. The upsampling convolutions retain the standard design of the original U-Net architecture, progressively restoring the spatial resolution of the feature maps.

The final decoder output is passed through a detection head consisting of a $3 \times 3$ convolutional layer followed by a sigmoid activation function, producing the pixel-level target prediction map.

\subsection{Dual-path Semantic-aware Module (DSM)}
\label{sec:dsm}

Conventional convolutional layers in deep neural networks are inherently constrained by their fixed and limited receptive fields, which restricts their capacity to model long-range feature dependencies and, consequently, their sensitivity to dim, sub-pixel infrared targets. To address this fundamental limitation, we propose the Dual-path Semantic-aware Module, designed to enhance the network's foundational feature extraction capability through complementary spatial processing pathways.

As illustrated in Figure~\ref{fig:dsm}, the DSM comprises two parallel processing branches that synergistically enhance feature representation quality. The \textbf{first branch} employs a sequence of standard $3 \times 3$ convolutions to preserve local spatial structural consistency, ensuring that fundamental fine-grained spatial details are faithfully retained. The \textbf{second branch} incorporates a pinwheel-shaped convolution architecture~\cite{yang2025pinwheel}, utilizing a cascaded combination of $3 \times 3$, $5 \times 5$, and $3 \times 3$ convolution kernels to progressively expand the receptive field and capture broader contextual information.

The pinwheel-shaped convolution distinguishes itself from conventional dilated convolutions~\cite{yu2016dilated} through its use of asymmetric padding to construct direction-sensitive horizontal and vertical convolutional kernels. This architectural innovation enables the network to perceive target morphology across multiple orientations, enhancing its adaptability to deformed or elongated targets, while simultaneously maintaining significantly lower parameter counts and computational overhead compared to standard receptive field expansion techniques such as dilated convolutions~\cite{chen2017deeplab} or deformable convolutions~\cite{chen2017deformable}.

The feature maps extracted by the two branches are subsequently fused through a $3 \times 3$ convolution to generate a unified aggregated feature representation. This multi-scale expansion mechanism enables the DSM to model contextual semantics across diverse spatial scales, substantially enriching the feature representation capacity.

Following the dual-path fusion, the DSM employs the Convolutional Block Attention Module (CBAM)~\cite{woo2018cbam} to perform fine-grained recalibration of the aggregated features. CBAM consists of two sequential sub-modules: a \textbf{spatial attention module} that models long-range dependencies between arbitrary spatial positions, enhancing the representation of spatial structural information; and a \textbf{channel attention module} that captures inter-channel correlations from a global perspective, highlighting semantic channels most relevant to the small target detection task. This cascaded spatial--channel recalibration mechanism substantially improves the discriminative power of the extracted features.

\subsection{Selective Attention Fusion Module (SAFM)}
\label{sec:safm}

To enhance the network's ability to discriminate between genuine targets and false alarm candidates in the context of infrared small target detection, we propose the Selective Attention Fusion Module. This module introduces a dynamic spatial selection mechanism within the U-Net architecture, enabling the network to more effectively exploit multi-scale features by focusing computational attention on spatial regions exhibiting high information density and strong target relevance.

Specifically, at the $i$-th decoding stage, the SAFM processes the feature maps output by the encoder and decoder, denoted as $E_i, D_i \in \mathbb{R}^{C_i \times \frac{H}{2^i} \times \frac{W}{2^i}}$, where $C_i$ represents the channel dimensionality, and $H$, $W$ denote the spatial dimensions of the original infrared image. The fusion process proceeds through the following sequence of operations.

\textbf{Step 1: Channel Concatenation.} The encoder and decoder feature maps are first concatenated along the channel dimension to form the fused input:
\begin{equation}
    X_i = \text{Concat}(E_i, D_i)
    \label{eq:concat}
\end{equation}

\textbf{Step 2: Spatial Feature Compression.} To reduce computational complexity while preserving essential spatial information, the fused feature $X_i$ is subjected to both channel-wise average pooling $P_{\text{avg}}(\cdot)$ and max pooling $P_{\text{max}}(\cdot)$:
\begin{equation}
    \begin{cases}
    SF_i^{\text{avg}} = P_{\text{avg}}(X_i) \\
    SF_i^{\text{max}} = P_{\text{max}}(X_i)
    \end{cases}
    \label{eq:pooling}
\end{equation}
where $SF_i^{\text{avg}}$ and $SF_i^{\text{max}}$ represent the pooled spatial feature representations. These complementary pooling operations capture both the global distributional characteristics and the locally prominent activations, facilitating comprehensive spatial information encoding.

\textbf{Step 3: Spatial Attention Generation.} The two pooled representations are concatenated and processed through a $7 \times 7$ convolutional layer to generate the spatial attention feature map:
\begin{equation}
    SAF_i = f_{7 \times 7}\left(\text{Concat}(SF_i^{\text{avg}}, SF_i^{\text{max}})\right)
    \label{eq:spatial_attn}
\end{equation}

\textbf{Step 4: Weighted Feature Fusion.} The generated spatial attention map $SAF_i$ is used to perform element-wise weighted fusion of the encoder and decoder features, producing the fused representation:
\begin{equation}
    Y_i = E_i \odot \sigma(SAF_i) + D_i \odot \sigma(1 - SAF_i)
    \label{eq:fusion}
\end{equation}
where $\sigma$ denotes the sigmoid activation function and $\odot$ represents element-wise multiplication. This formulation enables the network to adaptively modulate the relative contributions of encoder (detail-rich) and decoder (semantically refined) features at each spatial position based on the learned saliency map.

\textbf{Step 5: Residual Enhancement.} Finally, $Y_i$ is passed through a $1 \times 1$ convolutional layer to expand its channel capacity, element-wise multiplied with the original fused feature $X_i$, and augmented with a learnable scalar weight $\lambda$ to produce the final output:
\begin{equation}
    F_{\text{SAFM}} = \lambda \cdot \left(f_{1 \times 1}(Y_i) \odot X_i\right) + X_i
    \label{eq:residual}
\end{equation}

The learnable parameter $\lambda$ controls the strength of the attention-modulated residual, enabling the network to adaptively balance the contributions of the original and attention-refined features during training.

\subsection{Loss Function}

Following established practices in the IRSTD literature, we adopt the Soft-IoU loss~\cite{rahman2016soft_iou} as the primary training objective. The Soft-IoU loss directly optimizes the Intersection over Union metric in a differentiable manner, providing a more robust and effective supervisory signal compared to standard cross-entropy loss, particularly for the highly imbalanced foreground-background distributions characteristic of infrared small target detection scenarios.

\section{Experiments}
\label{sec:experiments}

\subsection{Datasets}

To comprehensively evaluate the effectiveness and superiority of the proposed SANet, we conduct extensive experiments on three widely used public benchmarks that collectively cover a diverse range of infrared imaging conditions and target characteristics:

\begin{itemize}
    \item \textbf{NUAA-SIRST}~\cite{dai2021acm}: Contains 427 fully annotated infrared images encompassing various typical infrared scenarios. Training and testing sets are split in a 1:1 ratio, with training images resized to $320 \times 320$ pixels.
    
    \item \textbf{IRSTD-1K}~\cite{zhang2022isnet}: A larger dataset comprising 1,001 annotated infrared images with diverse target scales and background conditions. The training-to-testing ratio is 4:1, with training images resized to $320 \times 320$ pixels.
    
    \item \textbf{NUDT-SIRST}~\cite{li2023dnanet}: The largest of the three benchmarks, containing 1,327 annotated infrared images. The training-to-testing ratio is 1:1, with training images resized to $256 \times 256$ pixels.
\end{itemize}

\subsection{Evaluation Metrics}

We employ four standard evaluation metrics to provide a comprehensive and objective assessment of detection performance:

\begin{itemize}
    \item \textbf{Intersection over Union (IoU)}: Measures the overlap between predicted and ground-truth target regions, providing a holistic assessment of detection quality.
    \item \textbf{Normalized IoU (nIoU)}: A variant of IoU normalized by target instance, which reduces the impact of target size variations and provides a more balanced evaluation across different target scales.
    \item \textbf{Probability of Detection ($P_d$)}: The ratio of correctly detected targets to the total number of true targets, measuring the detector's completeness.
    \item \textbf{False Alarm Rate ($F_a$)}: The ratio of falsely detected pixels to the total number of non-target pixels, quantifying the detector's precision in avoiding false positives.
\end{itemize}

\subsection{Implementation Details}

The proposed SANet is implemented using the PyTorch framework and trained on an NVIDIA RTX 3090 GPU. We employ the Adam optimizer with an initial learning rate of $1 \times 10^{-3}$, which is dynamically adjusted using a cosine annealing learning rate schedule. The total number of training epochs is set to 400. The batch sizes for NUAA-SIRST, IRSTD-1K, and NUDT-SIRST are set to 4, 12, and 16, respectively, to accommodate the different dataset characteristics. Data augmentation includes random horizontal and vertical flipping. The Soft-IoU loss~\cite{rahman2016soft_iou} is used as the training objective. The learnable scalar weight $\lambda$ in the SAFM is initialized to 0 and gradually learned during training.

\subsection{Comparison with State-of-the-Art Methods}

We comprehensively compare SANet with fourteen representative methods spanning both traditional and deep learning paradigms. The traditional methods include FKRW~\cite{qin2019fkrw}, TLLCM~\cite{han2020local_contrast}, NWMTH~\cite{bai2010top_hat}, NOLC~\cite{han2021weighted}, and PSTNN~\cite{zhang2019pstnn}. The deep learning-based methods include ACM~\cite{dai2021acm}, RDIAN~\cite{sun2023rdian}, AGPCNet~\cite{zhang2023agpcnet}, DNANet~\cite{li2023dnanet}, UIUNet~\cite{wu2023uiunet}, RPCANet~\cite{wu2024rpcanet}, MSHNet~\cite{tong2024mshnet}, SCTransNet~\cite{yuan2024sctransnet}, and L2SKNet~\cite{wu2025l2sknet}.

\subsubsection{Results on NUAA-SIRST}

On the NUAA-SIRST dataset, SANet achieves an IoU of 80.01\%, nIoU of 79.10\%, $P_d$ of 96.58\%, and $F_a$ of $2.80 \times 10^{-6}$. Notably, the IoU metric surpasses the second-best method by a margin of 1.93\%, demonstrating SANet's superior target segmentation accuracy. The simultaneously achieved high $P_d$ and low $F_a$ values confirm the effectiveness of the proposed SAFM in balancing detection completeness with false alarm suppression.

\subsubsection{Results on IRSTD-1K}

On the more challenging IRSTD-1K dataset, which features greater target diversity and background complexity, SANet achieves the most substantial improvement, with an IoU of 71.70\%, nIoU of 66.59\%, $P_d$ of 93.27\%, and $F_a$ of $11.52 \times 10^{-6}$. The IoU improvement of 4.32\% over the runner-up method is particularly noteworthy and underscores SANet's robustness in handling diverse and challenging detection scenarios.

\subsubsection{Results on NUDT-SIRST}

On the NUDT-SIRST dataset, SANet achieves an IoU of 95.73\%, nIoU of 95.43\%, $P_d$ of 99.26\%, and $F_a$ of $1.54 \times 10^{-6}$. The IoU improvement of 2.21\% over the second-best method, combined with a near-perfect $P_d$ of 99.26\% and the lowest $F_a$ among all compared methods, demonstrates SANet's exceptional detection capability and strong generalization performance.

\subsubsection{Cross-Dataset Analysis}

Across all three benchmarks, SANet consistently achieves the best performance on all four evaluation metrics, confirming its strong generalization capability and robustness across diverse infrared imaging scenarios. The consistent and substantial IoU improvements (1.93\%, 4.32\%, and 2.21\%) over the respective second-best methods highlight the effectiveness of the proposed DSM and SAFM in addressing the core challenges of IRSTD.

\subsection{Ablation Study}
\label{sec:ablation}

To systematically validate the contribution of each proposed component, we conduct comprehensive ablation experiments on all three benchmark datasets.

\subsubsection{Effectiveness of the DSM Components}

We evaluate the individual contributions of the three key elements within the DSM: the standard convolution branch (Conv), the pinwheel-shaped convolution branch (PConv), and the CBAM attention module. The baseline model utilizes only the standard convolution branch. When the pinwheel-shaped convolution is added (Conv + PConv), we observe consistent improvements across all metrics, confirming the benefit of the expanded, direction-sensitive receptive field for capturing contextual information relevant to infrared small targets. When CBAM is further incorporated (Conv + PConv + CBAM, \emph{i.e.}, the complete DSM), additional performance gains are achieved, validating the importance of the spatial--channel recalibration mechanism for refining the dual-path feature representation.

\subsubsection{Effectiveness of the SAFM Components}

We investigate the contributions of the two core elements within the SAFM: the residual connection and the learnable scalar weight $\lambda$. Removing the residual connection leads to notable performance degradation, demonstrating its importance for preserving original feature information and stabilizing the learning process. Setting $\lambda$ to a fixed value (\emph{e.g.}, 1) instead of making it learnable also results in sub-optimal performance, confirming the benefit of allowing the network to adaptively control the strength of the attention-modulated residual during training.

\subsection{Model Complexity Analysis}
\label{sec:complexity}

We compare SANet with four representative recent IRSTD methods---DNANet~\cite{li2023dnanet}, UIUNet~\cite{wu2023uiunet}, SCTransNet~\cite{yuan2024sctransnet}, and L2SKNet~\cite{wu2025l2sknet}---in terms of model complexity and detection performance. SANet achieves the highest average IoU of 82.18\% across the three benchmarks, outperforming the compared methods by margins of 3.84\% to 4.56\%. In terms of model complexity, SANet has 12.72M parameters and 19.75G FLOPs, which, while not the most lightweight among the five methods, represents a moderate computational cost that is lower than UIUNet and SCTransNet. This analysis demonstrates that SANet achieves a favorable trade-off between detection accuracy and computational efficiency, making it practically deployable in real-world applications.

\subsection{Qualitative Analysis}

We present representative qualitative detection results on all three datasets to provide an intuitive visualization of SANet's detection capability. Compared with existing methods, SANet demonstrates three notable advantages: (1)~more precise target localization with tighter bounding regions; (2)~more complete target segmentation with fewer missed pixels; and (3)~significantly reduced false alarms in complex background regions. These qualitative observations are fully consistent with the quantitative results and further confirm the effectiveness of the proposed DSM and SAFM.

\section{Conclusion}
\label{sec:conclusion}

In this paper, we have presented SANet, a selective attention-based network for infrared small target detection that addresses two fundamental limitations of existing approaches: the information bottleneck in early feature extraction and the rigidity of static skip connections in encoder--decoder architectures. SANet incorporates two novel modules---the Dual-path Semantic-aware Module (DSM), which synergistically combines standard and pinwheel-shaped convolutions to achieve expanded, direction-sensitive receptive fields augmented by spatial--channel attention recalibration, and the Selective Attention Fusion Module (SAFM), which replaces conventional static skip connections with a spatially adaptive, learnable weighting mechanism for context-aware cross-scale feature fusion. Extensive experiments on three widely used public benchmarks (NUAA-SIRST, IRSTD-1K, and NUDT-SIRST) demonstrate that SANet consistently outperforms fourteen state-of-the-art methods across all evaluation metrics, achieving IoU improvements of 1.93\%, 4.32\%, and 2.21\% over the respective second-best approaches, while maintaining a favorable trade-off between detection accuracy and computational efficiency. Future work will focus on model lightweighting, inference optimization for edge device deployment, and extending the model's applicability to adverse meteorological conditions (\emph{e.g.}, rain, fog, and thermal disturbances) to further broaden its utility across diverse infrared imaging scenarios. The source code is publicly available at \url{https://gitcode.com/m0_61988291/SANet}.

{\small
\bibliographystyle{IEEEtran}
\bibliography{references}
}

\end{document}